\documentclass[10pt,twocolumn,letterpaper]{article}

\usepackage{iccv}
\usepackage{times}
\usepackage{epsfig}
\usepackage{graphicx}
\usepackage{amsmath}
\usepackage{amssymb}
\usepackage{graphicx}
\usepackage{amsmath}
\usepackage{amssymb}
\usepackage{booktabs}

\usepackage{amsmath,amsfonts,bm}

\def\eqref#1{equation~\ref{#1}}

\def\1{\bm{1}}

\def\rmF{{\mathbf{F}}}

\def\vc{{\bm{c}}}
\def\vd{{\bm{d}}}

\def\vo{{\bm{o}}}

\def\vr{{\bm{r}}}

\def\vv{{\bm{v}}}

\DeclareMathAlphabet{\mathsfit}{\encodingdefault}{\sfdefault}{m}{sl}
\SetMathAlphabet{\mathsfit}{bold}{\encodingdefault}{\sfdefault}{bx}{n}

\def\sR{{\mathbb{R}}}

\usepackage{times}
\usepackage{epsfig}
\usepackage{epigraph}
\usepackage{graphicx}
\usepackage{amsmath}
\usepackage{amssymb}
\usepackage{tabularx}
\usepackage{multirow}
\usepackage{float}

\usepackage[table]{xcolor}
\usepackage{soul}
\usepackage{pifont}
\usepackage{wrapfig}
\usepackage{threeparttable}

\usepackage[pagebackref,breaklinks,colorlinks]{hyperref}

\iccvfinalcopy 

\def\iccvPaperID{6261} 

\usepackage[capitalize]{cleveref}
\crefname{section}{Sec.}{Secs.}
\Crefname{section}{Section}{Sections}
\Crefname{table}{Table}{Tables}
\crefname{table}{Tab.}{Tabs.}
\newcommand{\rev}[1]{#1}

\def\voxelname{general voxels\@\xspace}

\def\modelname{GNeuVox\@\xspace}
\def\pername{individual voxels\@\xspace}

\definecolor{best_color}{HTML}{FCE5CD}
\definecolor{better_color}{HTML}{DEEDF2}

\ificcvfinal\fi

\begin{document}

\title{Generalizable Neural Voxels for Fast Human Radiance Fields}


\author{
Taoran Yi$^{1}$\footnotemark[1] ,  Jiemin Fang$^{2,1}$\footnotemark[1] ,  Xinggang Wang$^1$\footnotemark[2] ,  Wenyu Liu$^1$\\
$^1$School of EIC, Huazhong University of Science \& Technology \\
$^2$Institute of Artificial Intelligence, Huazhong University of Science \& Technology \\
\texttt{\small\{taoranyi, jaminfong, xgwang, liuwy\}@hust.edu.cn}\\
}

\maketitle
{
\renewcommand{\thefootnote}{\fnsymbol{footnote}}
\footnotetext[1]{Equal contributions.}
\footnotetext[2]{Corresponding author.}}
\begin{abstract}
Rendering moving human bodies at free viewpoints only from a monocular video is quite a challenging problem. The information is too sparse to model complicated human body structures and motions from both view and pose dimensions. Neural radiance fields (NeRF) have shown great power in novel view synthesis and have been applied to human body rendering. However, most current NeRF-based methods bear huge costs for both training and rendering, which impedes the wide applications in real-life scenarios. In this paper, we propose a rendering framework that can learn moving human body structures extremely quickly from a monocular video. The framework is built by integrating both neural fields and neural voxels. Especially, a set of generalizable neural voxels are constructed. With pretrained on various human bodies, these general voxels represent a basic skeleton and can provide strong geometric priors. For the fine-tuning process, individual voxels are constructed for learning differential textures, complementary to general voxels. Thus learning a novel body can be further accelerated, taking only a few minutes. Our method shows significantly higher training efficiency compared with previous methods, while maintaining similar rendering quality. The project page is at \url{https://taoranyi.com/gneuvox}.
\end{abstract}

\section{Introduction}
\label{sec:intro}
Rendering human bodies~\cite{Wang_2021_CVPR,weng2020vid2actor,sanyal2021learning,habermann2021real,wang2021dance,wang2018video,sarkar2021style,neverova2018dense,ma2017pose,chan2019everybody,balakrishnan2018synthesizing} is a longstanding research topic and plays important roles in varieties of applications, \eg, virtual reality (VR), augmented reality (AR) and other interactive products. Recently, the emergence of neural radiance fields (NeRF)~\cite{mildenhall2020nerf} has significantly facilitated the development of rendering techniques. Taking sparse-view images as input, NeRF models can generate images from novel views with high qualities. 

Some works~\cite{peng2021neural,weng_humannerf_2022_cvpr,su2021anerf,chen2021animatable,peng2022animatable_expansion,peng2021animatable,chen2021snarf,jiang2022neuman,xu2021h} successfully apply NeRF methods to human body rendering frameworks. Benefiting from geometry priors, \eg SMPL models~\cite{SMPL:2015} for predicting human poses, body structures can be learned accurately even for poses with complicated motions. However, NeRF-based models rely on volume rendering to connect 2D image pixels with 3D real points, which requires for inferring tons of points and takes huge cost. The training process for one single scene usually needs dozens of hours or even days to complete on one GPU. This will undoubtedly impede these methods from applications in real-life scenarios.

In this paper, a fast training framework is built for rendering free-view images of human bodies from a monocular video recording a moving person. We first propose to represent the human body under the canonical pose with neural voxel features, which can be optimized via gradient descent directly. This explicit representation significantly accelerates the optimization process. Equipped with deformation networks that transform human poses into canonical ones, human body information can be modeled accurately and quickly.
More importantly, considering different human bodies share commonalities, \eg the basic skeleton, we enable the generalization ability of the proposed framework and name our method as \textbf{\modelname}. Two types of neural voxels are constructed to achieve this goal. One as \textbf{general voxels} is pretrained across various human bodies where common properties/information are learned and stored. The other one as \textbf{individual voxels} is built for any novel scene, which learns the scene's specific appearance and unique textures. With the collaboration of two types of neural voxels, the whole framework can learn a new human body extremely quickly and render high-quality images. 

We summarize contributions as follows.
\begin{itemize}
    \item A fast-training framework is designed by introducing explicit neural voxels, which renders free-view moving human bodies from a monocular video.
    \item As far as we know, we are the first to propose \textbf{generalizable explicit representations} to model common properties across different human bodies, which show significant effectiveness in accelerating novel human learning.
    \item The proposed framework is evaluated to show notably high efficiency with only \textbf{5-minute} training time. Even across different datasets, a strong generalization ability is still achieved. The code will be released.
\end{itemize}

\section{Related Work}
\label{sec: Related}

\subsection{Neural Rendering for Human Bodies}
Differentiable neural rendering has been widely adopted in human body scenarios which produces highly realistic images. Neural Body~\cite{peng2021neural} proposes to compute latent code volumes by inferring mesh vertices but performs poorly on unseen poses. To solve this problem, Animatable NeRF~\cite{peng2021animatable,peng2022animatable_expansion} maps human poses from the observation space to the predefined canonical space. Some other methods~\cite{Gao2022neuralnovelactor,liu2021neural} share similar ideas, which are not limited to the human body, but also apply to other dynamic scenes~\cite{pumarola2021d,park2021hypernerf,park2021nerfies,tineuvox,jiang2022instantavatar,instant_nvr}. In ARAH~\cite{ARAH:2022:ECCV} and SNARF~\cite{chen2021snarf}, a more accurate joint root-finding algorithm is used to obtain a more accurate mapping, which can better generalize to poses beyond the existing distribution. Neuman~\cite{jiang2022neuman} separates the background and characters, which are then modeled separately. There are also other methods to separate the scene~\cite{song2022nerfplayer,li2020neural}, which can lead to better rendering results. A-NeRF~\cite{su2021anerf} uses 3D skeleton poses to model the human body. In H-NeRF~\cite{xu2021h}, the combination of NeRF and Signed Distance Field (SDF) also obtains good human rendering results. PIFu~\cite{saito2019pifu} and PIHuHD~\cite{saito2020pifuhd} use the supervision of 3D mesh to model the human body.  ~\cite{chen2022gdna,bergman2022gnarf,wang2021metaavatar} achieve the effect of generating humans through additional supervision. ARCH~\cite{huang2020arch}, ARCH++~\cite{he2021arch++} and ICON~\cite{xiu2022icon} also require additional 3D supervision, but the cost of supervision data acquisition is expensive. Our method can achieve free-viewpoint rendering with only monocular videos, which greatly reduces the difficulty of dataset acquisition. Compared with other methods using monocular videos~\cite{li2020neural,huang2020arch,jiang2022selfrecon,alldieck2022phorhum,alldieck2018detailed,gao2021dynamic,tretschk2021non,gao_hang2022dynamic,weng2023personnerf}, ours also enjoys high convergence efficiency.

\subsection{Generalizable NeRF}
Generalization is an important but challenging problem in NeRF, as a NeRF model can usually only represent one specific scene. Making NeRF generalizable will greatly improve the efficiency of the representation. PixelNeRF~\cite{yu2021pixelnerf} projects images onto generalizable feature volumes. Inspired by MVSNet~\cite{yao2018mvsnet}, MVSNeRF~\cite{mvsnerf} builds a voxel feature containing relative positions and directions for view synthesis. These methods, including some follow-up methods~\cite{grf2020,mine2021,reizenstein21co3d,wang2021ibrnet}, extract features from images of views near the target view, so as to achieve the purpose of generalization. However, due to occlusion present in some poses, artifacts may occur in certain positions. For this reason, GeoNeRF~\cite{johari-et-al-2022} and NeuRay~\cite{liu2022neuray} gather information from consistent source views to reduce artifacts.
Some works also explore NeRF generalization on human bodies~\cite{Zhao_2022_CVPR,kwon2021neural,cheng2022generalizable,MPS_NeRF,KeypointNeRF_ECCV2022,chen2022geometry}, but the convergence speed is slow.

\subsection{NeRF Acceleration}
NeRF~\cite{mildenhall2020nerf} and its extensions~\cite{Barron_2021_ICCV,kaizhang2020,Martin_Brualla_2021_CVPR,huang2021hdr,barron2022mip,mildenhall2022nerf,ma2022deblur,lombardi2019neural,Wang_2021_CVPR} have achieved high rendering quality, but most of them bear long inference time and take large training cost to learn the scene. For inference speed, it can be promoted by improving 
the sampling strategy or baking related properties~\cite{Yu_2021_ICCV,Reiser_2021_ICCV,Hedman_2021_ICCV,Garbin_2021_ICCV,wang2022fourier}. DVGO~\cite{sun2021direct}, Plenoxels~\cite{yu_and_fridovichkeil2021plenoxels}, Instant-NGP~\cite{muller2022instant} and other methods~\cite{niessner2013real,sitzmann2019deepvoxels,xu2022point,reiser2023merf} significantly speed up the convergence speed by introducing explicit representations. DS-NeRF~\cite{kangle2021dsnerf} proposes to accelerate the convergence with depth supervision. TensoRF~\cite{Chen2022ECCV} uses tensor decomposition to achieve high storage efficiency while maintaining fast training speed. However, the aforementioned ones can only be applied to static scenes. \rev{\cite{tineuvox,liu2022devrf,gan2022v4d,peng2023intrinsicngp}} extend the efficient NeRF framework to dynamic scenes. We not only introduce explicit representations for acceleration but also make them generalizable for extremely fast convergence.

\section{Method}

\begin{figure*}[thbp]
\centering
\includegraphics[width=\linewidth]{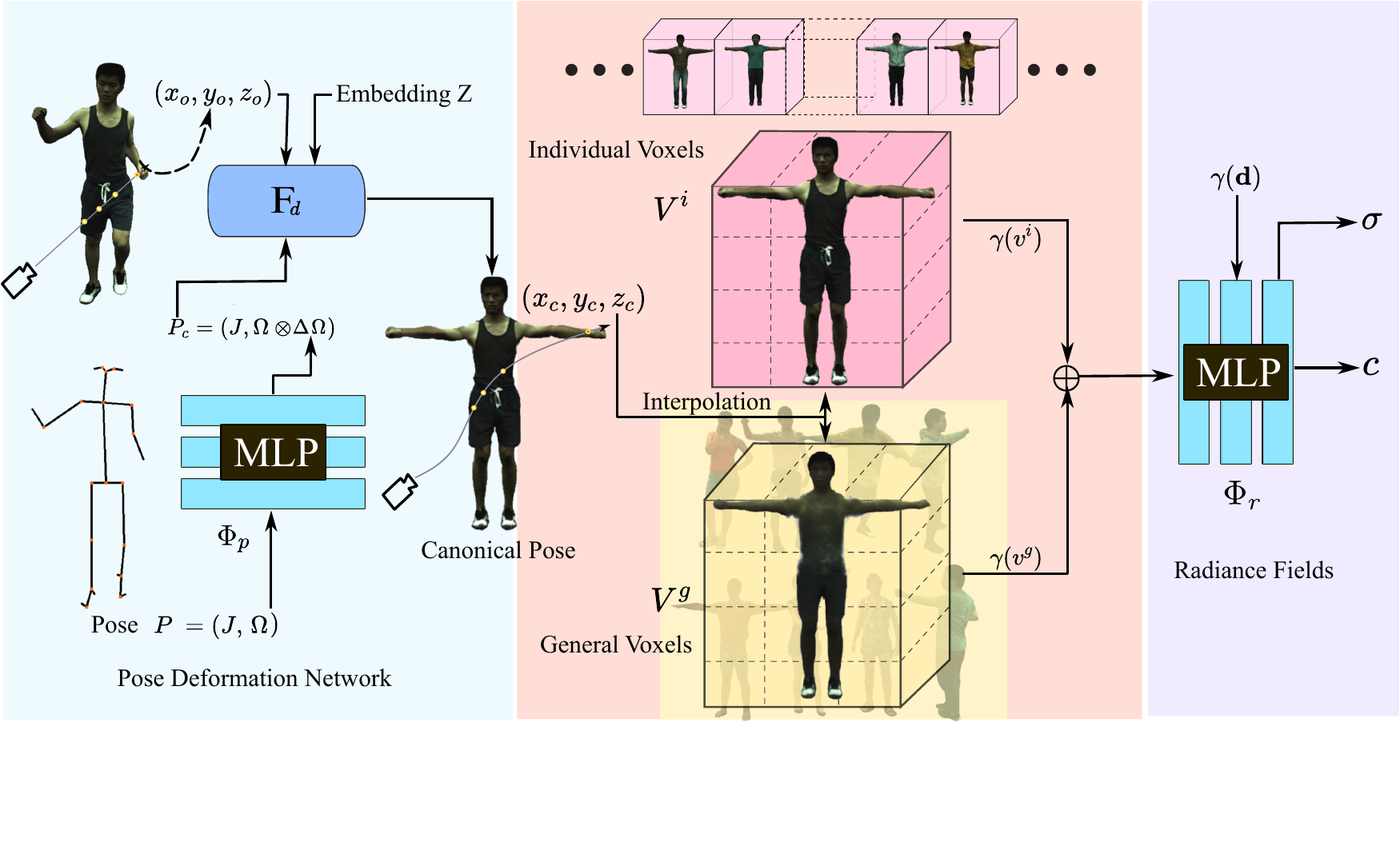}
\vspace{-58pt}
\caption{Overall framework of \modelname. First, we use the pose refinement module to adjust the obtained pose $P$, and then use the corrected pose $P_c$ and the embedding $Z$ to deform the coordinates $(x_o,y_o,z_o)$ of the human body in the observation space to the T-pose in the canonical space. Using the shifted coordinates $(x_c,y_c,z_c)$ to interpolate the individual voxels $V^{i}$ and general voxels $V^{g}$ to obtain their corresponding feature vectors $\vv^i$ and $\vv^g$. Finally, $\vv^i$ and $\vv^g$ are fed into the radiance fields to obtain the color $\vc$ and density $\sigma$.}
\label{fig: framework}
\vspace{-10pt}
\end{figure*}

In this section, we first review the original NeRF~\cite{mildenhall2020nerf} and HumanNeRF~\cite{weng_humannerf_2022_cvpr} methods in Sec.~\ref{subsec: pre}. Then in Sec.~\ref{subsec: Framework}, the proposed framework for fast-learning human bodies is introduced. We discuss how to generalize the framework across different scenes in Sec.~\ref{subsec: Generalize}. Finally, in Sec.~\ref{subsec: Optimization}, we elaborate the optimization process. 

\subsection{Preliminaries}
\label{subsec: pre}
Neural radiance fields (NeRF) are first proposed in \cite{mildenhall2020nerf}, aiming at connecting real 3D points with 2D pixels of images. NeRF is designed as a 5D function $f$, mapping 3D coordinates $(x,y,z)$ along with 2D directions $\vd = (\theta, \phi)$ into the color $\vc$ and density $\sigma$. This process can be formulated as:
\begin{equation}
   \begin{aligned}
      \label{eq: nerf}
      \vc , \sigma = f (x,y,z , \vd ),
   \end{aligned}
\end{equation}
where $f$ is commonly instantiated as multilayer perceptrons (MLPs).

In order to obtain the pixel color $C(\vr)$, points are sampled along the ray $\vr = \vo + t \vd$ emitted from the camera, where $\vo$ denotes the origin of the ray, and $\vd$ denotes the ray direction. Then the differentiable volume rendering~\cite{max1995optical} is performed to accumulate all 3D point colors and densities along the ray and produce the final pixel color:
\begin{equation}
   \begin{aligned}
      \hat{C}(\vr) &= \sum_{i=1}^N T_i(1 - \text{exp}(-\sigma_i \delta_i))\vc_i,
   \end{aligned}
\end{equation}
where $T_i = \text{exp}(- \sum_{j=1}^{i-1} \sigma_j  \delta_j)$, $\delta_i = t_{i+1} - t_{i}$ denotes the distance between two adjacent points, $N$ denotes the number of sampled points. To represent high-frequency details~\cite{tancik2020fourfeat}, $(x,y,z)$ and $\vd$ in Eq.~\ref{eq: nerf} are transformed with the positional encoding $\gamma$, which is formulated as
\begin{equation}
   \label{eq: pe}
   \gamma(x) =(sin(2^0x), cos(2^0x), ..., sin(2^{L-1}x), cos(2^{L-1}x)),
\end{equation}
where $L$ is used to control the maximum frequency.

Parameters of the field model are optimized to make the rendered color $\hat{C}(\vr)$ approach the ground truth pixel color ${C}(\vr)$. The loss function is defined as
\begin{equation}
   \label{eq: color_loss}
   \mathcal{L} = \lVert\hat{C}(\vr) - C(\vr)\rVert^2_2.
\end{equation}

HumanNeRF~\cite{weng_humannerf_2022_cvpr} is a representative work of extending NeRF to human bodies. It uses the NeRF model to represent a human body with the canonical pose. To achieve this, a pose refinement module and a coordinate deformation network are introduced to transform the human pose into the canonical one. The pose refinement module corrects the pose $P$ obtained from the off-the-shelf technique~\cite{SMPL:2015} to obtain a more accurate one $P_c$. The coordinate deformation network maps the observation pose $(x_o,y_o,z_o)$ to the canonical one $(x_c, y_c, z_c)$ with the corrected pose.
The human body pose can be represented as $P = (J, \Omega)$, $J$ denotes the positions of $K$ human body key points and $\Omega$ denotes the rotation angles of local joints. The pose refinement network is denoted as $\Delta \Omega = \Phi_{p}(\Omega)$, which is performed to get a more
accurate pose $P_c=(J_c,\Omega_c )$. It can be formulated as
\begin{equation}
   \begin{aligned}
   \label{eq: pose correction}
      P_c=(J_c,\Omega_c ) = (J, \Omega \otimes \Delta \Omega ),
   \end{aligned}
\end{equation}
where $\otimes $ denotes $(R^i \cdot \Delta  R^i, T^i + \Delta T^i)$, $R^i$ and $T^i$ denote the rotation matrix and translation matrix to the canonical space corresponding to the i-th bone. $(R^i,T^i)$ and $(\Delta R^i,\Delta T^i)$ are derived from $\Omega$ and $\Delta \Omega$ respectively. 

The coordinate deformation network~\footnote{The deformation network in HumanNeRF~\cite{weng_humannerf_2022_cvpr} contains two parts of rigid skeletal offsets and non-rigid offsets. Here we only refer to skeletal offsets for simplicity.} is formulated as:
\begin{equation}
\begin{aligned}
\label{eq: skeletal offsets}
    (x_{skel}, y_{skel},z_{skel}) = T_{skel}((x_o,y_o,z_o) , P_c),\\
    T_{skel}(x , P_c) =\sum_{i=1}^{K}{w{^i}(x)(R_c{^i} x + T_c{^i})},
\end{aligned}
\end{equation}
where $(x_o,y_o,z_o)$ is denoted as $x$ for abbreviation and  $(R_c{^i},T_c{^i})$ is derived from $\Omega_c$, $w{^i}$ is a weight parameter that controls the blend weight for the i-th bone. $w{^i}$ is obtained by mapping a learnable embedding vector $Z$ into a volume via several 3D convolutional layers, which is further trilinearly interpolated. 

\subsection{Fast-training Framework for Rendering Human Bodies}
\label{subsec: Framework}
Our framework is constructed with the following modules, \ie the pose deformation network, neural voxels, and radiance fields. We show the overall framework of our method in Fig.~\ref{fig: framework}. 

\vspace{-10pt}
\paragraph{Overall Pipeline}
We first use the pose refinement module to correct the obtained pose $P$, and use the corrected pose $P_c$ with the embedding $Z$ to deform the coordinates $(x_o, y_o, z_o)$ of the human body in the observation space to the T-pose in the canonical space through the pose deformation network $\rmF_d$. Then we use the deformed coordinates $(x_c,y_c,z_c)$ in the canonical space to interpolate the neural voxels $V$ to get the voxel feature $\vv$. Finally, $\vv$ together with the coordinates $(x_o,y_o,z_o)$, direction $\vd$, and timestamp $t$ are fed into the radiation field to obtain the color $\vc$ and density $\sigma$.

\vspace{-12pt}
\paragraph{Pose Deformation Network}
We construct the pose refinement network and coordinate deformation network, following HumanNeRF~\cite{weng_humannerf_2022_cvpr}, to model human motions. The process of mapping the pose in the observation space $(x_o,y_o,z_o)$ to the canonical space $(x_c,y_c,z_c)$ is formulated as
\begin{equation}
\begin{aligned}
\label{eq: defom}
    (x_c,y_c,z_c) = \rmF_d((x_o,y_o,z_o), P_c),
\end{aligned}
\end{equation}
where $P_c$ is the human body pose after pose refinement network $\Phi_{p}$, which is the same as Eq.\ref{eq: pose correction}. $\rmF_d$ has the same structure as $T_{skel}$ in Eq.\ref{eq: skeletal offsets}, which is achieved via 3D convolutional layers with corrected poses $P_c$ and zero-initialized embedding $Z$.


\vspace{-12pt}
\paragraph{Representing Canonical Bodies with Neural Voxels}
Most previous NeRF-based methods for human bodies~\cite{peng2021neural,weng_humannerf_2022_cvpr,su2021anerf,chen2021animatable,peng2022animatable_expansion,peng2021animatable,chen2021snarf,jiang2022neuman,xu2021h} adopt purely implicit representations. Although good rendering quality is obtained, it takes a long time to complete the training phase. To speed up the training on human bodies, an explicit data structure is introduced, namely \emph{neural voxels}. We construct a set of optimizable feature parameters which are organized into a voxel grid structure. These voxel features are designed for representing the canonical T-pose human body, which ease the difficulty of optimization.
Then trilinear interpolation is performed on \emph{neural voxels} $V\in \sR^{C_v \times N_x \times N_y \times N_z}$ to get the features $\vv$, which is performed as multi-distance interpolation (MDI)~\cite{tineuvox}. $C_v$ denotes the channel number of each voxel feature, while $N_x$, $N_y$, and $N_z$ denote the length of each dimension in 3D space. 
\begin{equation}
   \begin{aligned}
   \label{eq: MDI}
      \vv = \text{interp} \{ (x_c,y_c,z_c) ,V\} ,\\
   \end{aligned}
\end{equation}

where $\text{interp}$ denotes trilinear interpolation operation.
Besides, the timestamp $t$, direction $\vd$, original coordinates $(x_o,y_o,z_o)$ along with interpolated voxel features $\vv$ are all fed into the radiance field $\Phi_{r}$ to predict the final color $\vc$ and density $\sigma$.
\begin{equation}
   \begin{aligned}
   \label{eq: radiance_net}
      \vc,\sigma = \Phi_r(\gamma(\vv),\gamma(t), \gamma(x_o, y_o, z_o), \gamma(\vd)),
   \end{aligned}
\end{equation}
where $\gamma$ is the positional encoding function as defined in Eq.~\ref{eq: pe}.

\renewcommand{\arraystretch}{1.20}
\begin{table*}[htbp]
\setlength{\tabcolsep}{1pt}
\centering
\caption{Quantitative comparisons on the ZJU-MoCap dataset~\cite{peng2021neural}. GNeuVox represents training from scratch. GNeuVox-ZJU and GNeuVox-H36m represent the fine-tuning results after pretraining on the ZJU-MoCap and Human3.6M~\cite{ionescu2013human3} dataset. We color cells with the best values in \colorbox{best_color}{orange} and the second best values in \colorbox{better_color}{blue}.}
\vspace{0pt}
\begin{tabular}{c | c | c | c | c | c | c | c | c | c}
\hline
\multirow{2}{*}{} &  \multicolumn{3}{c|}{Subject 377} & \multicolumn{3}{c|}{Subject 386} & \multicolumn{3}{c}{Subject 387} \\ 
\cline{2-10}
 & PSNR $\uparrow$ & SSIM $\uparrow$ & LPIPS ($\times 10^{-2}$) $\downarrow$ & PSNR $\uparrow$ & SSIM $\uparrow$ & LPIPS ($\times 10^{-2}$) $\downarrow $ & PSNR $\uparrow$ & SSIM $\uparrow$ & LPIPS ($\times 10^{-2}$) $\downarrow$ \\ 
\hline
Neural Body~\cite{peng2021neural} & 29.11 & 0.9674 & 4.095 & 30.54 & 0.9678 & 4.643 & 27.00 & 0.9518 & 5.947 \\
\hline
HumanNeRF~\cite{weng_humannerf_2022_cvpr} & \cellcolor{best_color}30.41  & 0.9743 & \cellcolor{best_color}2.406 & \cellcolor{better_color}33.20 & 0.9752 & 2.899 & 28.18 & \cellcolor{best_color}0.9632 & \cellcolor{best_color}3.558 \\

\hline
\modelname & 29.77 & 0.9754 & 2.673 & 33.17  &  0.9755  &  3.070  & 28.05 & 0.9595  &4.270 \\
\hline
\modelname-ZJU  & 29.93& \cellcolor{best_color}0.9763 & \cellcolor{better_color}2.408& \cellcolor{best_color}33.45& \cellcolor{best_color}0.9769 &\cellcolor{best_color}2.771  & \cellcolor{best_color}28.43 & \cellcolor{better_color}0.9623 &\cellcolor{better_color}3.790\\
\hline
\modelname-H36m & \cellcolor{better_color}30.03 & \cellcolor{better_color}0.9761 & 2.500 & 33.02  & \cellcolor{better_color}0.9758 & \cellcolor{better_color}2.840 & \cellcolor{better_color}28.24  & 0.9613 &4.028\\
\hline
\multirow{2}{*}{} &  \multicolumn{3}{c|}{Subject 392} & \multicolumn{3}{c|}{Subject 393} & \multicolumn{3}{c}{Subject 394} \\ 
\cline{2-10}
 & PSNR $\uparrow$ & SSIM $\uparrow$ & LPIPS ($\times 10^{-2}$) $\downarrow$ & PSNR $\uparrow$ & SSIM $\uparrow$ & LPIPS ($\times 10^{-2}$) $\downarrow $ & PSNR $\uparrow$ & SSIM $\uparrow$ & LPIPS ($\times 10^{-2}$) $\downarrow$ \\ 
\hline
Neural Body~\cite{peng2021neural} & 30.10 & 0.9642 & 5.327 & \cellcolor{best_color}28.61 & 0.9590 & 5.905 & 29.10 & 0.9593 & 5.455\\
\hline
HumanNeRF~\cite{weng_humannerf_2022_cvpr} & \cellcolor{better_color}31.04 & \cellcolor{best_color}0.9705 & \cellcolor{best_color}3.212 & 28.31 & \cellcolor{better_color}0.9603 & \cellcolor{best_color}3.672 & \cellcolor{better_color}30.31 & \cellcolor{better_color}0.9642 & \cellcolor{best_color}3.289\\

\hline
\modelname & 30.81 & 0.9687 & 3.536 & \cellcolor{better_color}28.53  &  0.9594  &   4.072 & 29.83  & 0.9620  & 3.822\\

\hline
\modelname-ZJU    & \cellcolor{best_color}31.07& \cellcolor{best_color}0.9705 &\cellcolor{better_color}3.376 & \cellcolor{better_color}28.53& \cellcolor{best_color}0.9612 & \cellcolor{better_color}3.794 & \cellcolor{best_color}30.49 & \cellcolor{best_color}0.9648 & \cellcolor{better_color}3.439 \\

\hline
\modelname-H36m  & 30.78 & 0.9694 & 3.469& 28.38&  0.9604 & 3.962 &  30.21& 0.9635  &3.593\\
\hline
\end{tabular}
\label{tab: all scene}
\end{table*}

\begin{table*}[thbp]
   \newcommand{\xmark}{\ding{55}}
   \centering
   \setlength{\tabcolsep}{6pt}
   \caption{Comparisons about training cost and rendering quality on ZJU-MoCap dataset~\cite{peng2021neural}. We list the number of training iterations on each scene and the corresponding training time. We color cells with the best values in \colorbox{best_color}{orange} and the second best values in \colorbox{better_color}{blue}.}
   \vspace{0pt}
   \begin{tabular}{l|ccc|ccc}
   \toprule
   \textbf{Method} & \textbf{Pretrain Dataset} & \textbf{Perscene Iterations}&\textbf{Time} & \textbf{PSNR} $\uparrow$ & \textbf{SSIM} $\uparrow$ & \textbf{LPIPS} 
    ($\times 10^{-2}$) $\downarrow$   \\
   \midrule
   Neural Body~\cite{peng2021neural}  & \xmark & -- & -- &29.08 & 0.9616 &  5.229 \\
   HumanNeRF~\cite{weng_humannerf_2022_cvpr} & \xmark & 400k & 288 hours  &30.24 & \cellcolor{better_color}0.9680 &  \cellcolor{best_color}3.173  \\
   \modelname & \xmark & 10k& 50 mins &\cellcolor{best_color}30.26& 0.9678
   & 3.450
   \\
   \modelname & ZJU-MoCap & 1k & \cellcolor{best_color}5 mins&\cellcolor{best_color}30.26 & \cellcolor{best_color}0.9682 & 3.420  \\
   \modelname & Human3.6M~\cite{ionescu2013human3} & 3k & \cellcolor{better_color}15 mins&30.11 &0.9677  &\cellcolor{better_color}3.399   \\
   \bottomrule
   \end{tabular}
   \label{tab: mean metric}
   \vspace{-12pt}
   \end{table*}
   

\subsection{Generalization across Human Bodies}
\label{subsec: Generalize}
Rendering human bodies is a domain-specific task. Though different human bodies own various appearances or wear diverse clothes, they share commonalities in the basic skeleton. To take full advantage of the geometry priors, we propose to enable the generalization ability of the framework. This is achieved by constructing two types of neural voxels introduced in Sec.~\ref{subsec: Framework}. One is designed as \textbf{general voxels} $V^{g}$ which are pretrained across various human bodies. General voxels extract substantial information from different human bodies, representing a basic skeleton or template applicable for any human body. The other one is constructed instantly as \textbf{individual voxels} $ V^{i}$, which are optimized to represent unique appearances for a specific human body. The interpolation process,  which is same as Eq.~\ref{eq: MDI}, is performed on both voxel grids to obtain voxel features $\vv^{g}$ and $\vv^i$, coming from $V^{g}$ and $ V^{i}$ respectively. Two voxel features $\vv^{g}$ and $\vv^i$ are concatenated and fed into the radiance network. The radiance field function in Eq.~\ref{eq: radiance_net} is re-defined as: $\vc,\sigma = \Phi_r(\gamma(\vv^g), \gamma(\vv_n^i),\gamma(t), \gamma(x_o, y_o, z_o), \gamma(\vd))$. With pretrained general voxels, optimization of the rendering framework directly starts from a body template, not from nothing or chaos. Then the training phase can be further accelerated, benefiting from structure priors obtained from various human bodies.

The training process is divided into two phases, \ie one pretraining phase across multiple different human bodies and fine-tuning for a specific human. During the pretraining, besides general voxels, the deformation network and radiance field are also shared for all humans. The two networks can be treated as decoders and the decoding function becomes stronger and more accurate with seeing substantially different humans. On the other hand, the individual voxels and deformation embedding $Z$ are built separately for each specific human. 
For the fine-tuning phase, a new set of individual voxels and embedding $Z$ are built with a zero initialization. Parameters of all modules in the framework are updated to fit the new human body. The fine-tuning phase can be completed with quite few iterations, where mainly differential textures need to be captured.


\subsection{Optimization}
\label{subsec: Optimization}

The mean square error (MSE) between the rendered color and the ground truth is used as one term of our loss function as defined in Eq.~\ref{eq: color_loss}. Following HumanNeRF~\cite{weng_humannerf_2022_cvpr}, we add the learned perceptual image patch similarity (LPIPS)~\cite{zhang2018unreasonable} loss with the VGG~\cite{simonyan2014very} network to improve rendering details. Our final loss function can be formulated as
\begin{equation}
   \label{eq: our_loss}
   \mathcal{L} = \lambda_{M}\mathcal{L}_{MSE} + \lambda_{L}\mathcal{L}_{LPIPS},
\end{equation}
where $\lambda_{M}$ and $\lambda_{L}$ control the magnitudes of two loss terms respectively.

In each iteration, we sample rays of $G$ patches with a size of $H \times H$ on one image for computing the LPIPS loss. To speed up training, we only use the MSE loss in the first $I_n$ iterations during fine-tuning. In the left iterations, the LPIPS loss is added to improve the details.

\section{Experiments}
\begin{figure*}[thbp]
   \centering
   \includegraphics[width=\linewidth]{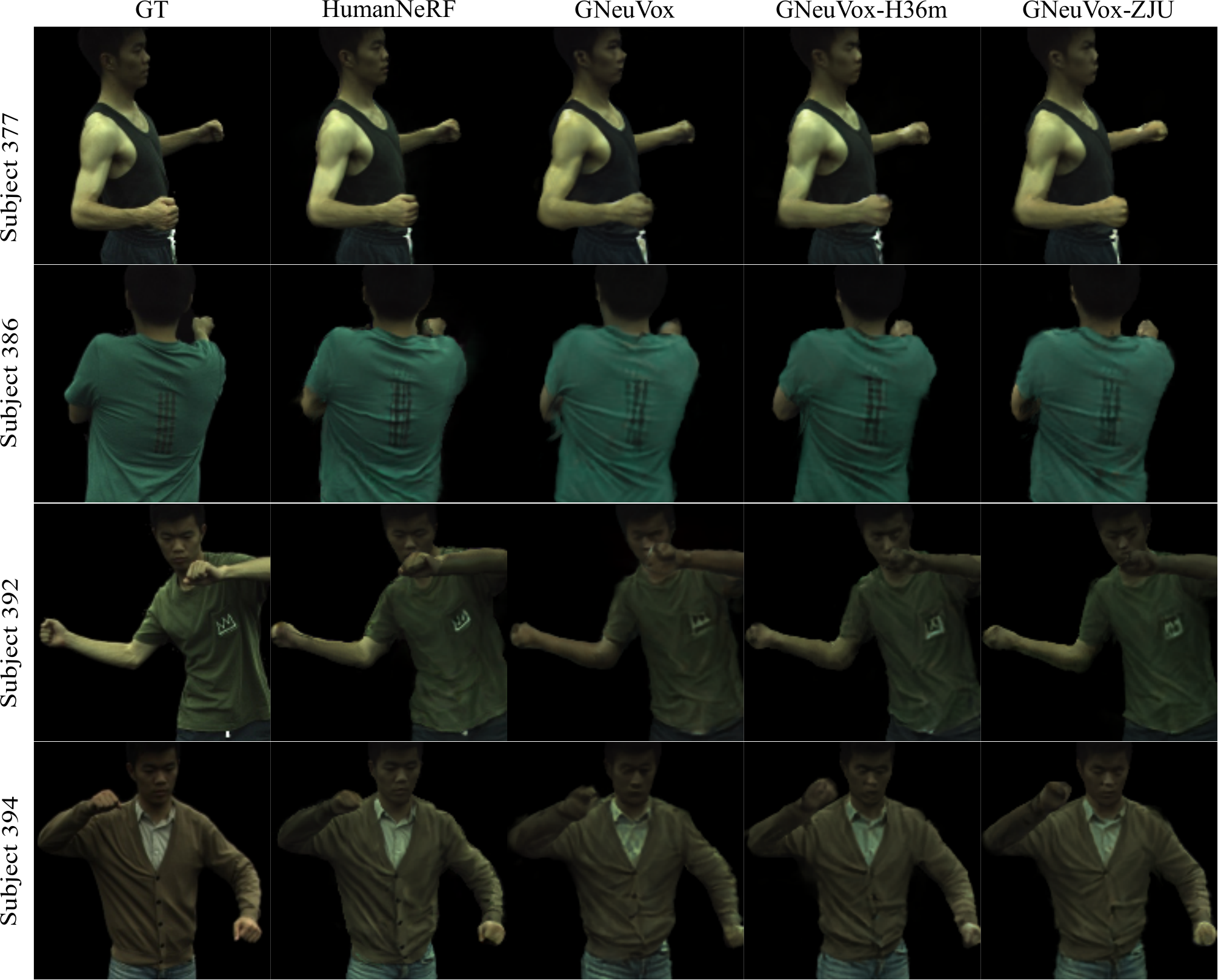}
   \vspace{-10pt}
   \caption{Qualitative comparisons between HumanNeRF~\cite{Zhao_2022_CVPR} and our GNeuVox on the ZJU-MoCap~\cite{peng2021neural} dataset. GNeuVox represents training from scratch, GNeuVox-ZJU and GNeuVox-H36m represent the fine-tuning results with pretraining on the ZJU-MoCap and Human3.6M~\cite{ionescu2013human3} dataset respectively.}
   \label{fig: vis_all}
   \vspace{-12pt}
\end{figure*}

In this section, we first describe the implementation details in Sec.~\ref{subsec: implem}.
Then we show evaluation results and compare with other related methods in Sec.~\ref{subsec: eval}.
In Sec.~\ref{subsec: abla}, ablation studies and a series of analysis are performed for better understanding the key components of \modelname.
\subsection{Implementation Details}
\label{subsec: implem}

We implemented our framework using PyTorch~\cite{paszke2019pytorch}. The \voxelname are constructed at a resolution of $160^3$. We build \pername separately for each scene with the same resolution as \voxelname. The channel number of features in both types of voxel grids is set as 6. For voxel features $\vv$, we use positional encoding with frequencies of 2. Frequencies of positional encoding on the coordinates $(x,y,z)$ and direction $\vd$ are set as 10 and 4 respectively. We provide more details of the pose deformation network and radiation network in supplementary materials.

For training, we use an Adam~\cite{kingad2015methodforstochasticoptimization} optimizer with $(0.9,0.99)$ $\beta$ values.
Our training is divided into two phases, pretraining and fine-tuning.
During both stages, we initialize the learning rate as $5\times 10^{-5}$, except voxel grids and $\Phi_{r}$ are set as $2\times 10^{-2}$ and $5\times 10^{-4}$ respectively. The learning rate decays by a factor of 0.1 for every $500k$ iterations. The pretraining phase takes $50k$ iterations in total. The model is fine-tuned for $1k$ iterations when pretrained on the ZJU-MoCap dataset~\cite{peng2021neural} and for $3k$ iterations when pretrained on Human3.6M~\cite{ionescu2013human3}. We also provide a training-from-scratch version for $10k$ iterations. All experiments are performed on one single GeForce RTX 3090 GPU.

In practice, we get 6 32$\times$32 patches for sampling rays. 128 points are randomly sampled along each ray. For pretraining or training from scratch, we set $\lambda_{M}$ as 0.2 and $\lambda_{L}$ as 1 in Eq.\ref{eq: our_loss}. $\lambda_{M}$ is set as 10 and $\lambda_{L}$ as 0 in the first 300 iterations during fine-tuning. For the rest iterations, $\lambda_{M}$ is set as 0.2 and $\lambda_{L}$ as 1.

\begin{figure*}[t]
\centering
\includegraphics[width=\linewidth]{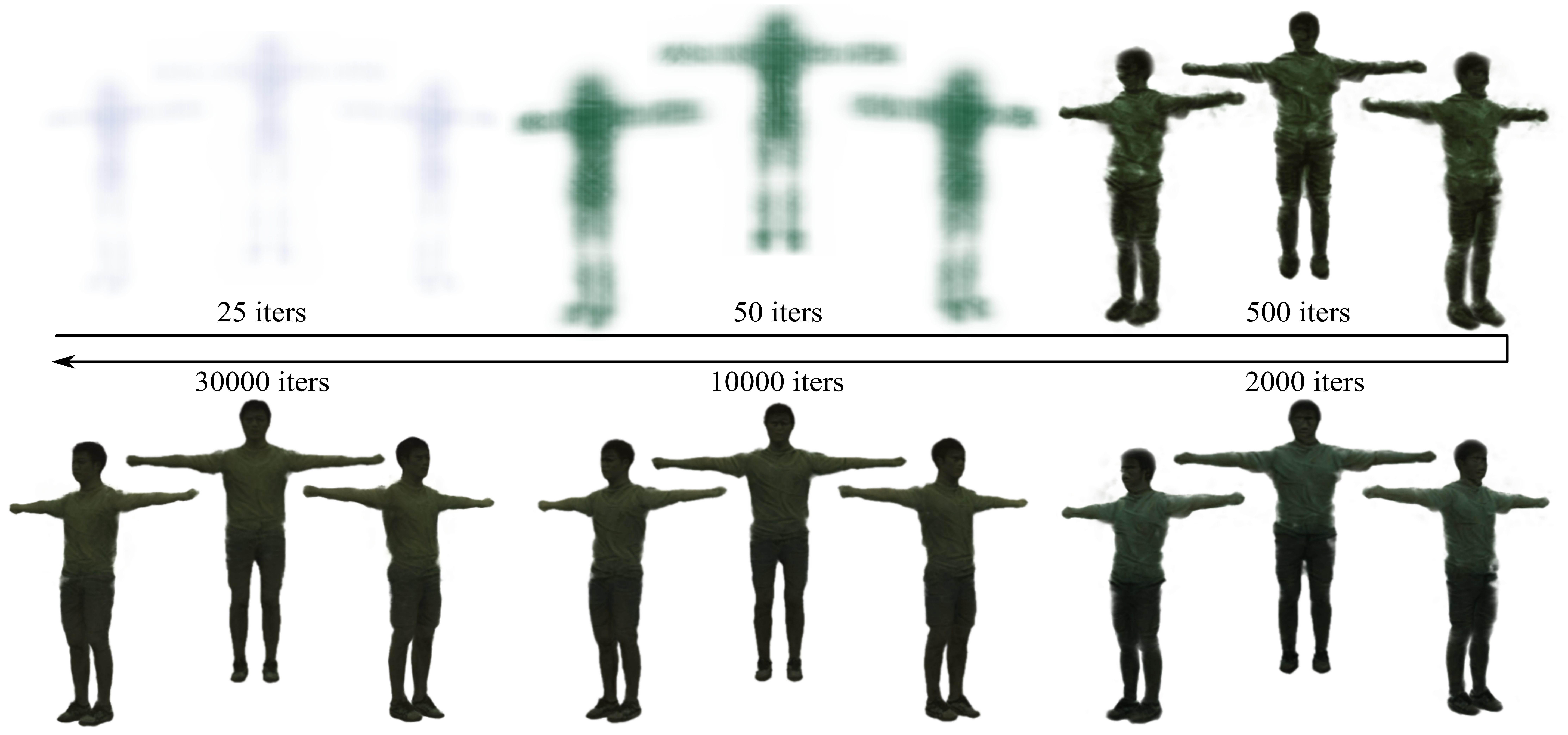}
\vspace{-10pt}
\caption{Change process of \voxelname during pretraining on the ZJU-MoCap~\cite{peng2021neural} dataset.}
\label{fig: genvoxel}
\vspace{-6pt}
\end{figure*}

\subsection{Evaluation}
\label{subsec: eval}
\paragraph{Training from Scratch }
When training from scratch, We only use the images taken by ``\textit{camera 1}'' in ZJU-MoCap~\cite{peng2021neural} as training for the monocular setting. And the remaining 22 camera views, except for ``\textit{camera 1}'' are used as evaluation. The selected images ensure that the human body rotates 360 degrees. We downsample the image to the half resolution of 540$\times$540. These settings are consistent with HumanNeRF~\cite{weng_humannerf_2022_cvpr}. The comparison results of our method, Neural body~\cite{peng2021neural}, and HumanNeRF are shown in Tab.~\ref{tab: all scene}.
We use three metrics for evaluation, namely peak signal-to-noise ratio (PSNR), structural similarity (SSIM), and LPIPS~\cite{zhang2018unreasonable}. 
The rendering results from Neural body are worse than other methods, as it is designed for the multi-view setting. 
HumanNeRF and our method can achieve similar performance on the one-view setting. Neural body and HumanNeRF take days to converge. In Tab.~\ref{tab: mean metric}, our method trained from scratch can get comparable results on ZJU-MoCap in less than 50 minutes. Here we only show the evaluation results for 6 subjects.

\paragraph{Pretraining on ZJU-MoCap}
We adopt the leave-one-out evaluation approach for ZJU-MoCap pertaining, which is also used in MPS-NeRF~\cite{MPS_NeRF}. 7 out of 8 subjects (313, 377, 386, 387, 390, 392, 393, 394) are selected for pretraining while the left one is used for fine-tuning and evaluation, so we need to pretrain more than once on ZJU-MoCap to get pretrained models. This manner guarantees no overlap between datasets for pertaining and fine-tuning. When evaluation, datasets are no different from training from scratch. As shown in Tab.~\ref{tab: mean metric}, training time can be shortened to 5 minutes if we load a pretrained model. 

\begin{table}[t!]
\newcommand{\xmark}{\ding{55}}
\centering
\caption{Comparisons about rendering quality on PeopleSnapshot~\cite{alldieck2018detailed}. GNeuVox-ZJU and GNeuVox-H36m represent the fine-tuning results after pretraining on the ZJU-MoCap~\cite{peng2021neural} and Human3.6M~\cite{ionescu2013human3} dataset.}
\vspace{0pt}
\setlength{\tabcolsep}{8pt}
\begin{tabular}{l|c|c}
\toprule
Method & Time & PSNR  $\uparrow$ \\
\midrule
Neural Body~\cite{peng2021neural} & $\sim$14 hours  &24.47 \\
\midrule
Anim-NeRF~\cite{chen2021animatable} & $\sim$13 hours   & 28.89 \\
\midrule
GNeuVox-ZJU & 5 mins & 27.73  \\
\midrule
GNeuVox-H36m & 5 mins  &27.94  \\
\bottomrule
\end{tabular}
\label{tab: PeopleSnapshot}
\vspace{-12pt}
\end{table}

\paragraph{Pretraining on Human3.6M~\cite{ionescu2013human3}}
For pretraining on Human3.6M, 7 subjects (S1, S5, S6, S7, S8, S9, and S11) preprocessed by Animatable NeRF~\cite{chen2021animatable} are used. Then the pretrained model is evaluated across datasets on ZJU-MoCap. For each subject in Human3.6M, all four views are used. Images are sampled for every five frames. The image is downsampled to half the original size to 500$\times$500. Note that the pretrain process only needs to be performed once on Human3.6M and applied to all subjects in ZJU-MoCap. Tab.~\ref{tab: mean metric} shows it only takes 15 minutes to converge with a Human3.6M-pretrained model and the PSNR is comparable to HumanNeRF. It is observed that the quantitative results have big variances when the model is trained from scratch. And sometimes the training may go collapse. Not only with our method, but the same problem also occurs with HumanNeRF. However, with pretrained general neural voxels and networks, the above situation is alleviated, and more stable training results can be achieved. The prior knowledge stored in \voxelname can help our method to fit stably on new human bodies.

\begin{table}[t!]
\newcommand{\xmark}{\ding{55}}
\centering
\caption{Ablation study about the generalization effects of several key components. }
\vspace{0pt}
\setlength{\tabcolsep}{1pt}
\begin{tabular}{ccc|ccc}
\toprule
    \begin{tabular}{c}\textbf{General}\\\textbf{Voxels} \end{tabular}
    & \begin{tabular}{c} \textbf{Radiance}\\ \textbf{Fields} \end{tabular} & 
   \begin{tabular}{c} \textbf{Deform}\\\textbf{CNN}\end{tabular} 
   &\textbf{PSNR}  $\uparrow$ & \textbf{SSIM} $\uparrow$ & \textbf{LPIPS} $\downarrow$ \\
\midrule
\checkmark & \checkmark & \checkmark &30.17 & 0.9678 & 3.725 \\
\midrule
\xmark& \checkmark & \checkmark & 30.04& 0.9663& 4.493 \\
\midrule
\checkmark &  \xmark &\checkmark &29.44 & 0.9625 & 4.955  \\
\midrule
\checkmark & \checkmark &\xmark & 30.14& 0.9669 &4.064 \\
\midrule
\xmark & \xmark &\xmark &  28.89 & 0.9540 & 6.561\\
\bottomrule
\end{tabular}
\label{tab: abla-gen-comps}
\vspace{-12pt}
\end{table}

\paragraph{Evaluation on PeopleSnapshot~\cite{alldieck2018detailed}}
To explore the model's ability to synthesize free-viewpoint of loose clothing, we conduct experiments on PeopleSnapshot using the same setting as ~\cite{chen2021animatable}. 
Due to the different human body pose definitions used, we recalculate the poses using ~\cite{easymocap}. 
We show the average PSNR of four scenes in in Tab.~\ref{tab: PeopleSnapshot}, and the results of Neural Body~\cite{peng2021neural} and Anim-NeRF~\cite{chen2021animatable} are from Anim-NeRF.
Our method can achieve results comparable to the current state-of-the-art method (Anim-NeRF) in just 5 minutes by loading the pretrained model. Even on human bodies wearing loose clothing, our method converges quickly, demonstrating strong generalization performance. Our method has indeed learned a good human template in \voxelname, leading to fast convergence.

\paragraph{Results of Visualization}
In Fig.~\ref{fig: vis_all} we show rendered images of training from scratch for 3k iterations and fine-tuning the same iterations for each scene.
It can be observed that the rendering results are significantly improved with the pretrained model, which are comparable to HumanNeRF with far less training cost. Pretraining on the same dataset and across datasets can both speed up the convergence, especially in terms of better facial expressions and clothing details.

\subsection{Ablation Study and Analysis}
\label{subsec: abla}
\paragraph{Generalization Effects of Separate Components}
We selectively and partially load the pretrained model to explore the generalization effectiveness of several components in the fine-tuning phase.
We conduct several experiments, and at each time we choose not to load a part of the pretrained network and fine-tune the model for 500 iterations. Three evaluated parts of the network include the convolutional layers in $\rmF_d$, \voxelname $V^g$ and the radiance field $\Phi_r$. 
As shown in Tab.~\ref{tab: abla-gen-comps}, the three components all play important roles for short-time training. Promotions brought by \voxelname and $\Phi_r$ are particularly prominent, 0.13dB and 0.73dB PSNR respectively.
When loading the pretrained \voxelname, the template features stored in it assist \pername to learn the unique features of the scene, and the two together represent the information of the entire human. Loading the pretrained radiation field can well decode the features in the two types of voxels. The pretrained CNN in $\rmF_d$ can help quickly find out how to convert the coordinates of the observation space to the canonical space at the beginning of training. Making all these components generalizable guarantees the best acceleration performance.

\begin{figure}[t!]
   \centering
   \vspace{-8pt}
   \includegraphics[width=\linewidth]{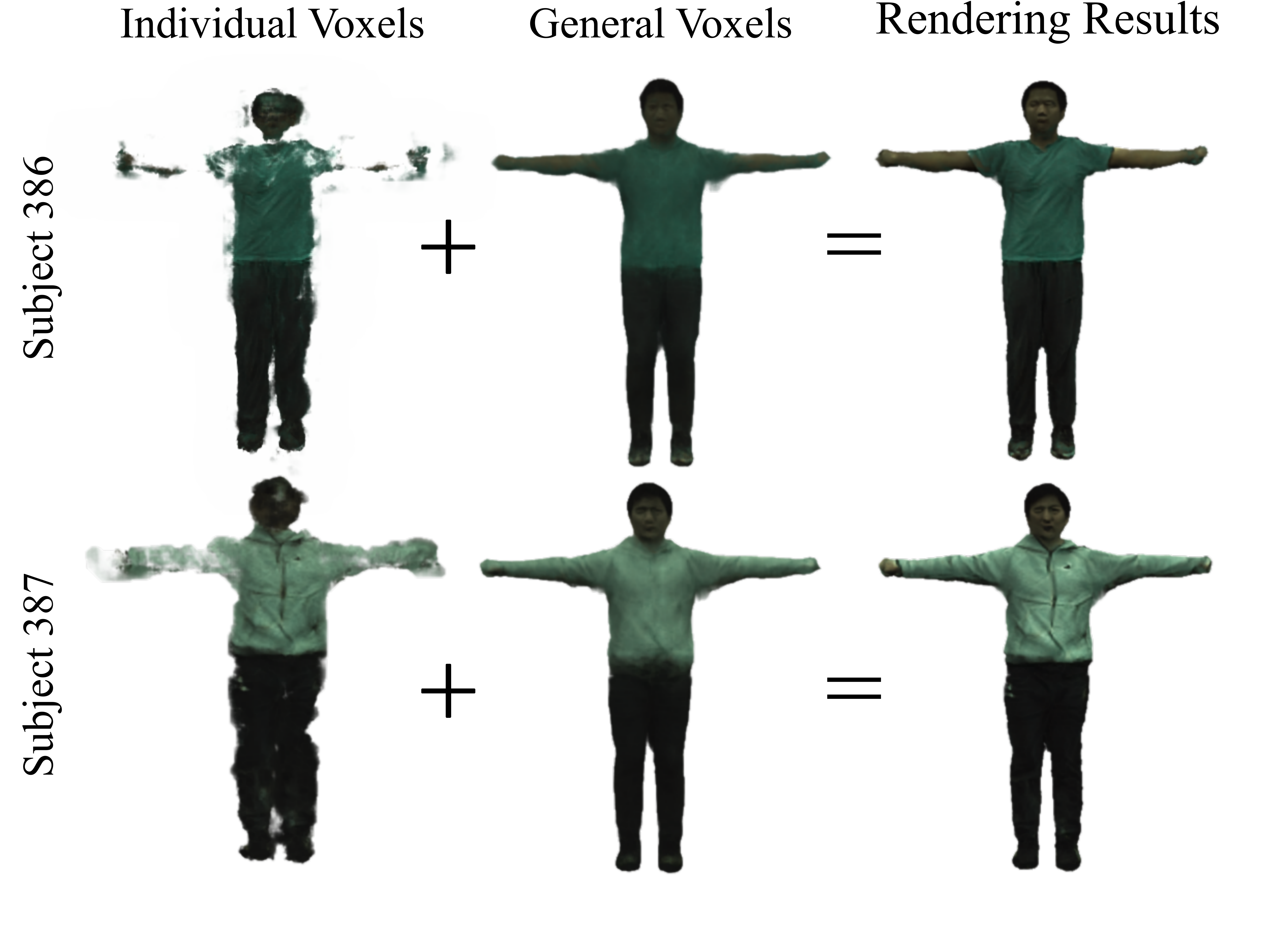}
    \vspace{-25pt}
   \caption{Rendering results by inferring general voxels and individual voxels separately.}
   \label{fig: voxels_vis}
    \vspace{-12pt}
\end{figure}

\paragraph{Changes of General Voxels in Pretraining}
In Fig.~\ref{fig: genvoxel}, we show the change process of \voxelname in the pretraining process.
It can be observed that \voxelname first gradually learn the outline of the human body in T-pose, but they learn some detailed information such as clothes folds in the middle pretraining process. This fold information is specific and different for each human. In the later pretraining process, these contents are degraded and the learned human becomes more general.
Besides, the approximate positions of facial features are gradually learned. This information is beneficial for \pername, at the fine-tuning phase, to fill the details of facial features into \voxelname, which results in rapid convergence.

\paragraph{Roles of Two Neural Voxels}
We analyze how the learned features of both voxels look like after fine-tuning. To achieve this, we set the parameters of the two voxels to zero, respectively, and obtain the result of rendering the two voxels separately after fine-tuning. The rendering result of \pername is very similar to the texture map, including details such as clothes folds and facial expressions. Although it consists of feature vectors that can be optimized, it exhibits obvious physical meaning. The rendered result of \voxelname supplements the approximate shapes and colors of the clothes, compared to that before fine-tuning as in Fig.~\ref{fig: voxels_vis}. The final rendering result is obtained by combining the differential textures contained in \pername and the contour features of \voxelname.

\section{Discussion}
\paragraph{Limitations}
It is observed that human face rendering shows vagueness and distortion to some degree. We analyze the reason as follow. Geometry priors used in the proposed generalization framework mainly comes from the human pose (SMPL) and skeleton (general voxels). As the fine-tuning iterations are quite few ($\le$ 3k), it is quite hard to learn the face contour accurately which may include complex non-rigid deformation. Besides, the clothes wrinkles do not match the ground truth image exactly. Improving the method towards better non-rigid deformation will be an important and interesting direction.

\paragraph{Future Works \& Potential}
Besides the aforementioned orthogonal direction, we would like to further demonstrate the potential value our method carries. The generalization ability of the proposed general neural voxels is proved to be effective on human bodies in this paper. We plan to extend the pretraining process to some large-scale human body datasets~\cite{tao2021function4d,cai2022humman} and explore its effects on learning downstream human structures. More importantly, we believe the general voxels can be a promising data structure for representing scenes/objects, not limited to human bodies. General neural voxels, as \textbf{generalizable explicit representations}, enjoy high efficiency in optimization while they are more explainable than implicit representations. It is expected that general voxels can be applied to more categories of real-life objects, \eg vehicles, animals and plants \etc.

\section{Conclusion}
In this paper, we build a fast-training framework for rendering free-view moving human bodies from a monocular video. Neural voxels are introduced to represent humans and accelerate the optimization phase. Besides, we propose to construct two types of neural voxels, one pretrained on various human bodies to extract a basic skeleton while the other one targeted at a specific human. With two neural voxels collaborating, the training phase can be completed in an extremely short time. The concept of general voxels, enjoying high optimization efficiency, is expected to be extended for modeling more categories of objects.

{\small
\bibliographystyle{ieee_fullname}
\bibliography{egbib}
}

\appendix
\section{Appendix}

\begin{table}[t!]
\newcommand{\xmark}{\ding{55}}
\centering
\caption{Ablation study about the number of pretraining iterations on fine-tuning results.}
\begin{tabular}{c|ccc}
\toprule
	\begin{tabular}{c}\textbf{Pretrain}\\\textbf{Iterations} \end{tabular} 
	&\textbf{PSNR}  $\uparrow$ & \textbf{SSIM} $\uparrow$ & \textbf{LPIPS} ($\times 10^{-2}$)  $\downarrow$ \\
\midrule
50k & 30.11 &0.9677 & 3.399 \\
200k & 30.26 &0.9681 &3.489 \\
\bottomrule
\end{tabular}
\label{tab: iter-pretrain}
\end{table}

\begin{table}[t!]
\newcommand{\xmark}{\ding{55}}
\centering
\caption{Ablation study about fine-tuning iterations with the pretrained model on the ZJU-MoCap dataset~\cite{peng2021neural}.}
\vspace{0pt}
\setlength{\tabcolsep}{15pt}
    \begin{tabular}{c|c|c|c}
    \toprule
    \textbf{Iterations} & 500  & 1000 & 3000 \\
    \midrule
    \textbf{PSNR}  $\uparrow$ &30.17 & 30.26 & 30.32 \\
    \bottomrule
    \end{tabular}
\label{tab: abla-iters}
\end{table}
\begin{figure}[t!]
\centering
\includegraphics[width=\linewidth]{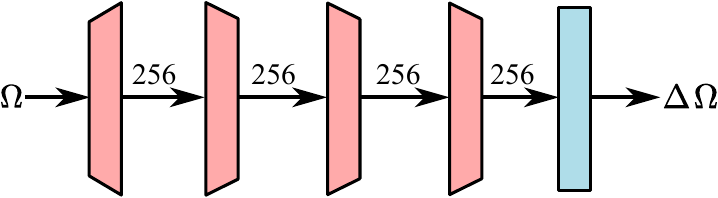}
\caption{Architecture of the pose refinement network $\Phi_{p}$. The red trapezoidal squares denote that ReLU activation function is applied, and the blue square denotes no activation function.}
\label{fig: pose}
\end{figure}

\begin{figure}[t!]
\centering
\includegraphics[width=\linewidth]{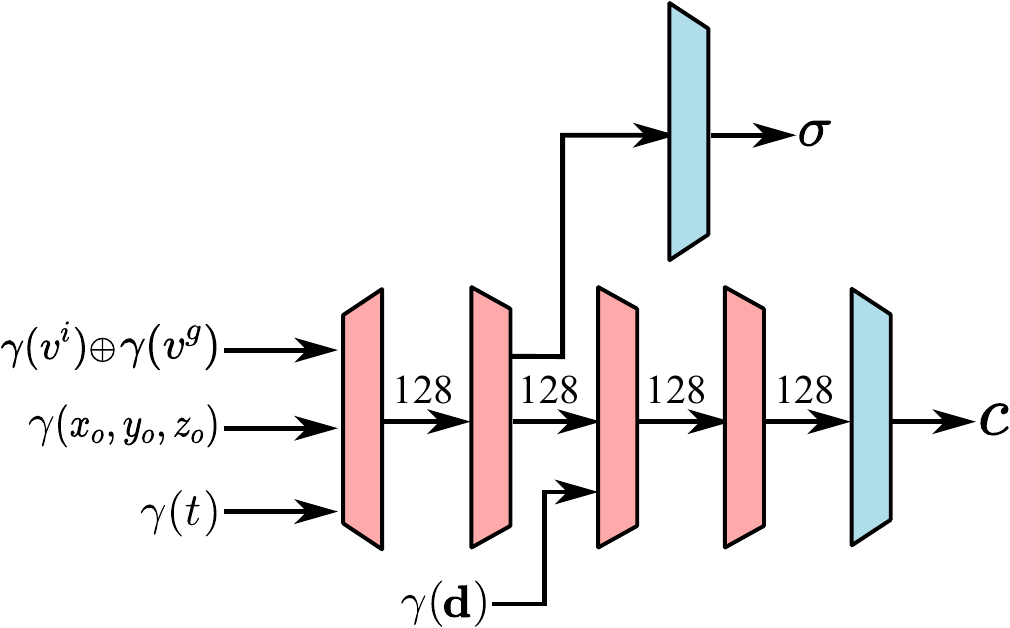}
\caption{Architecture of the radiance network $\Phi_{r}$. The blue trapezoidal squares denote that other activation functions are applied. The activation functions applied to the density $\sigma$ and color $\vc$ are softplus~\cite{Barron_2021_ICCV,sun2021direct} and sigmoid respectively.  $\gamma$ denotes the positional encoding.}
\label{fig: radiance}
\vspace{-5pt}
\end{figure}

\begin{figure}[t!]
\centering
\includegraphics[width=\linewidth]{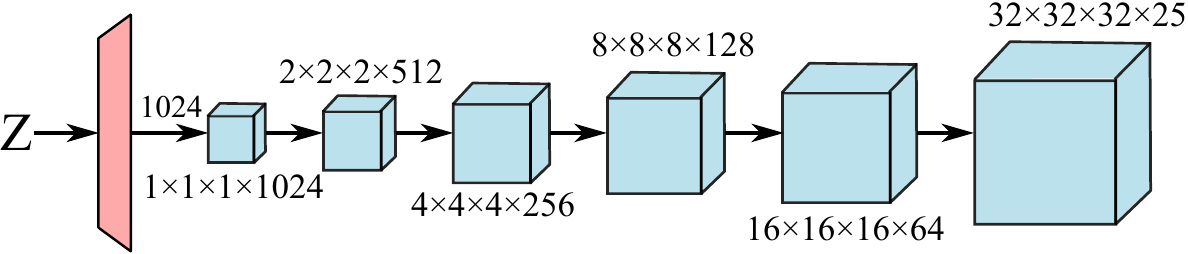}
\caption{The convolutional layers in the pose deformation network. First, we use one layer MLP to reshape the embedding $Z$ to 1$\times$1$\times$1$\times$1024, and then generate a volume of 32$\times$32$\times$32$\times$25 by transposed convolutions.}
\label{fig: cnn}
\end{figure}

\begin{figure*}[t]
\centering
\includegraphics[width=\linewidth]{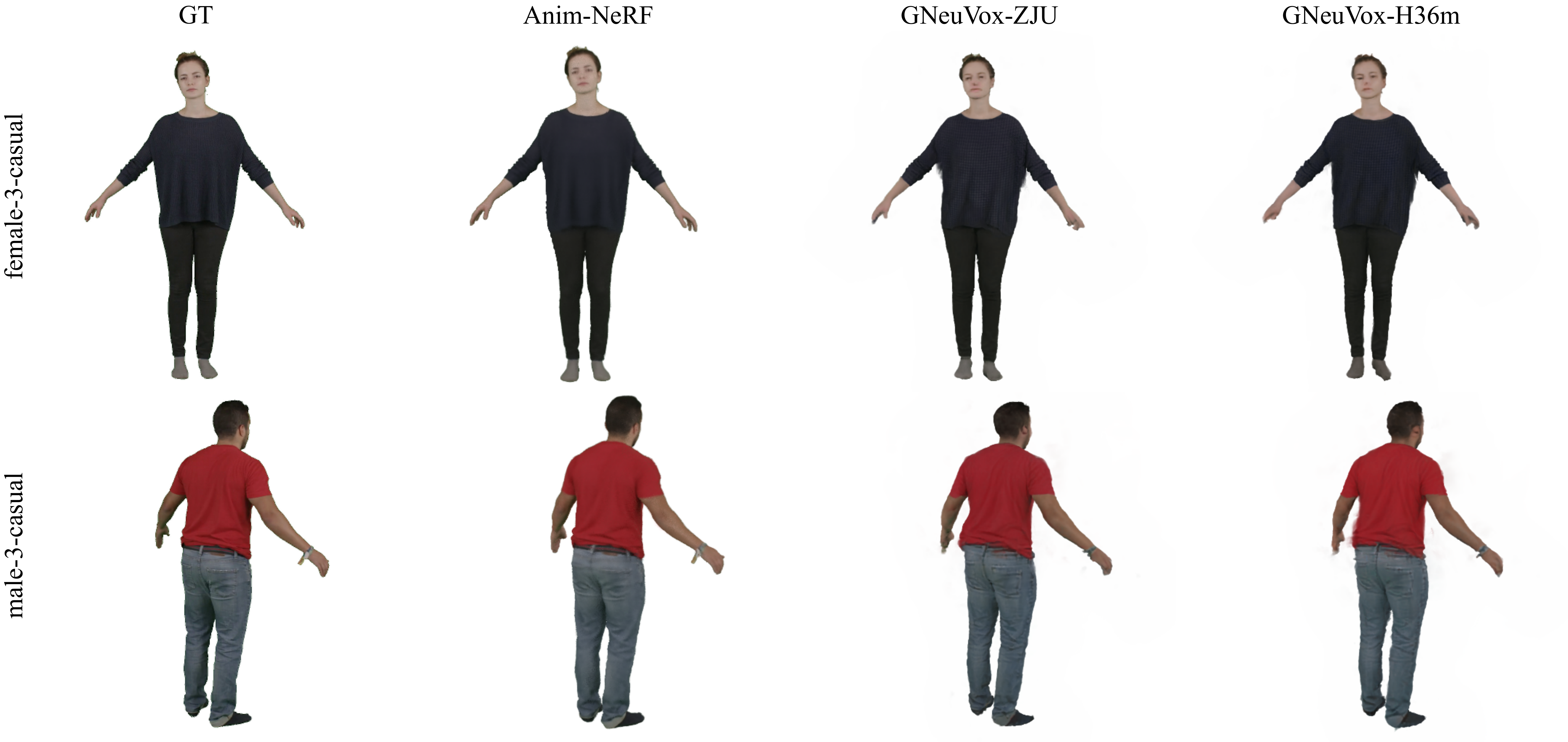}
\caption{Qualitative comparisons between Anim-NeRF~\cite{Zhao_2022_CVPR} and our GNeuVox on the PeopleSnapshot~\cite{alldieck2018detailed} dataset. GNeuVox-ZJU and GNeuVox-H36m denote the fine-tuning results with pretraining on the ZJU-MoCap~\cite{peng2021neural} and Human3.6M~\cite{ionescu2013human3} datasets respectively.}
\label{fig: snap}
\end{figure*}

\subsection{Neural Architectures}
In this section, we provide detailed structures of the pose refinement network $\Phi_{p}$, the convolutional layers in the pose deformation network $\rmF_d$, and the radiance network $\Phi_{r}$.
As shown in Fig.~\ref{fig: pose} and Fig.~\ref{fig: radiance}, $\Phi_{p} $  and $\Phi_{r} $ are both instantiated as MLPs. Fig.~\ref{fig: cnn} shows that through one layer MLP and transposed convolutions, the embedding $Z$ is transformed into a volume, which is trilinearly interpolated to obtain the blend weigh $w{^i}$ in Eq.\ref{eq: skeletal offsets} for the coordinates deformation.

\subsection{More Ablation Studies}
We pretrain the model on Human3.6M~\cite{ionescu2013human3} for different iterations and then fine-tune it on the ZJU-MoCap~\cite{peng2021neural} dataset for $3k$ iterations to explore the impact of pretraining iterations on fine-tuning results. As shown in Tab.~\ref{tab: iter-pretrain}, more iterations on the pretraining stage show little effect on fine-tuning results. 
The Human3.6M dataset has only 7 human bodies, and $50k$ iterations of pretraining are enough. We might get better fine-tuning results if we pretrain on more human bodies.

Moreover, we fine-tune the model with different iterations on ZJU-MoCap. Based on the experiment results as shown in Tab.~\ref{tab: abla-iters}, more fine-tuning iterations do not lead to a significant improvement in rendering quality.


\subsection{More Qualitative Comparisons}
We conduct experiments on the PeopleSnapshot~\cite{alldieck2018detailed} dataset, where human bodies are wearing looser clothing. We show the visualization results in Fig.~\ref{fig: snap}. Despite our method not constructing specialized modules for modeling non-rigid clothing or facial details, GNeuVox still achieves decent results. We leave building stronger modules for non-rigid deformations as an important future direction.

\end{document}